%% file: acl_latex.tex
\documentclass[11pt]{article}
\PassOptionsToPackage{dvipsnames,table,xcdraw}{xcolor}
\usepackage{xcolor}
\usepackage[final]{acl}
\usepackage{multirow}
\usepackage{pifont}
\usepackage{booktabs}
\usepackage{times}
\usepackage{latexsym}
\usepackage{adjustbox}
\usepackage[most]{tcolorbox}
\usepackage[normalem]{ulem}
\useunder{\uline}{\ul}{}
\newtcolorbox{prompt}[1]{
    enhanced,
    left=4mm,
    right=4mm,
    top=2mm,
    bottom=2mm,
    boxsep=0mm,
    rounded corners,
    title=#1,
    fontupper=\footnotesize\linespread{0.9}\fontfamily{lmr}\selectfont,
    }

\usepackage[T1]{fontenc}

\usepackage[utf8]{inputenc}

\usepackage{microtype}

\usepackage{inconsolata}

\usepackage{graphicx}
\usepackage{algorithm}
\usepackage{algorithmic}
\usepackage{setspace}
\usepackage{hyperref}
\usepackage{enumitem}

\definecolor{myyellow}{HTML}{fff9eb} 
\definecolor{myred}{HTML}{fbe7e6} 
\definecolor{mygreen}{HTML}{eaf3ea} 
\definecolor{paperblue}{HTML}{077dea}

\title{\textit{ToolSpectrum }: Towards Personalized Tool Utilization \\ for Large Language Models}

\author{Zihao Cheng$^{1,3\dagger}$ \thanks{$^\dagger$ Equal Contributions. Work done during the first author's internship at Beihang University.}, Hongru Wang$^{4\dagger}$ , Zeming Liu$^{1\ddagger}$  \thanks{$^\ddagger$ Corresponding Author}, Yuhang Guo$^{2\ddagger}$,\\ \bf   Yuanfang Guo$^{1}$, Yunhong Wang$^{1}$, Haifeng Wang$^{5}$\\
  $^{1}$School of Computer Science and Engineering, Beihang University, Beijing, China, \\
  $^{2}$School of Computer Science and Technology, Beijing Institute of Technology, Beijing, \\
  $^{3}$University of Science and Technology Beijing, Beijing, China, \\
  $^{4}$The Chinese University of Hong Kong, Hong Kong, China,   
  $^{5}$ Baidu Inc., Beijing, China \\
  \texttt{chengzihao008@gmail.com, hrwang@se.cuhk.edu.hk, zmliu@buaa.edu.cn} }\date{}
\makeatletter
\def\thanks#1{\protected@xdef\@thanks{\@thanks
        \protect\footnotetext{#1}}}
\makeatother

\begin{document}
\maketitle

\input{sections/1_abstract}
\input{sections/2_introduction}

\input{sections/3_related}

\input{sections/4_dataset}

\input{sections/5_experiment}

\input{sections/6_analysis}

\section{Conclusion}
This paper introduces personalized tool utilization, which considers functionally similar tools and user-specific factors to optimize tool utilization in real-world scenarios. Specifically, we define two critical factors: user profile and environment, and present \texttt{ToolSpectrum}, a benchmark designed to evaluate personalized tool utilization. Through extensive experiments conducted on \texttt{ToolSpectrum}, we demonstrate that incorporating personalized tool utilization significantly enhances its effectiveness. However, current LLMs face challenges in performing effectively on this new task.

\section*{Limitations}

This paper introduces \texttt{ToolSpectrum}, a benchmark designed to assess LLMs' performance in personalized tool utilization. However, a major limitation of this evaluation is the excessive context length required. This arises from the need to include detailed descriptions of Apps, APIs, parameters, and the necessary profile and environment data. The inclusion of such extensive information significantly increases the context length, which challenges the model’s ability to manage it effectively. As a result, this imposes a heavy computational burden, reducing the efficiency and effectiveness of the performance evaluation.

\section*{Ethics Statement}
In developing \texttt{ToolSpectrum}, we adhere strictly to established ethical standards, ensuring full compliance with legal and regulatory requirements throughout the data collection and processing stages. \texttt{ToolSpectrum} has been meticulously curated to exclude any content that promotes violence, discrimination, hate speech, or other harmful behaviors. We construct \texttt{ToolSpectrum} using carefully selected seed data, allowing for both control over its composition and transparency. This approach ensures we can effectively identify and mitigate biases related to race, gender, ethnicity, age, and other sensitive attributes. Our commitment to fairness, inclusivity, and equity is central to the design and evaluation processes. To further ensure accountability, we conduct regular ethical reviews to address potential risks and societal impacts, upholding the highest standards of transparency throughout.

\section*{Acknowledgements}
Thanks for the insightful comments and feedback from the reviewers. This work was supported by the National Key R\&D Program of China (No. 2023YFF0725600), the National Natural Science Foundation of China (No. 62406015), and CCF-Baidu Open Fund (No. CCF-BAIDU202411).

\bibliography{custom}
\input{sections/7_appendix}

\end{document}

%% file: sections/1_abstract.tex
\begin{abstract}

While integrating external tools into large language models (LLMs) enhances their ability to access real-time information and domain-specific services, existing approaches focus narrowly on functional tool selection following user instructions, overlooking the context-aware personalization in tool selection. This oversight leads to suboptimal user satisfaction and inefficient tool utilization, particularly when overlapping toolsets require nuanced selection based on contextual factors. To bridge this gap, we introduce \textbf{\texttt{ToolSpectrum}}, a benchmark designed to evaluate LLMs’ capabilities in personalized tool utilization. Specifically, we formalize two key dimensions of personalization, \textit{user profile} and \textit{environmental factors},  and analyze their individual and synergistic impacts on tool utilization. Through extensive experiments on \texttt{ToolSpectrum}, we demonstrate that personalized tool utilization significantly improves user experience across diverse scenarios. However, even state-of-the-art LLMs exhibit the limited ability to reason jointly about user profiles and environmental factors, often prioritizing one dimension at the expense of the other. Our findings underscore the necessity of context-aware personalization in tool-augmented LLMs and reveal critical limitations in current models. Our data and code are available at \url{https://github.com/BUAA-IRIP-LLM/ToolSpectrum}.

\end{abstract}

%% file: sections/2_introduction.tex
\section{Introduction}
\begin{figure}[t!]
    \centering
    \includegraphics[width=0.5\textwidth]{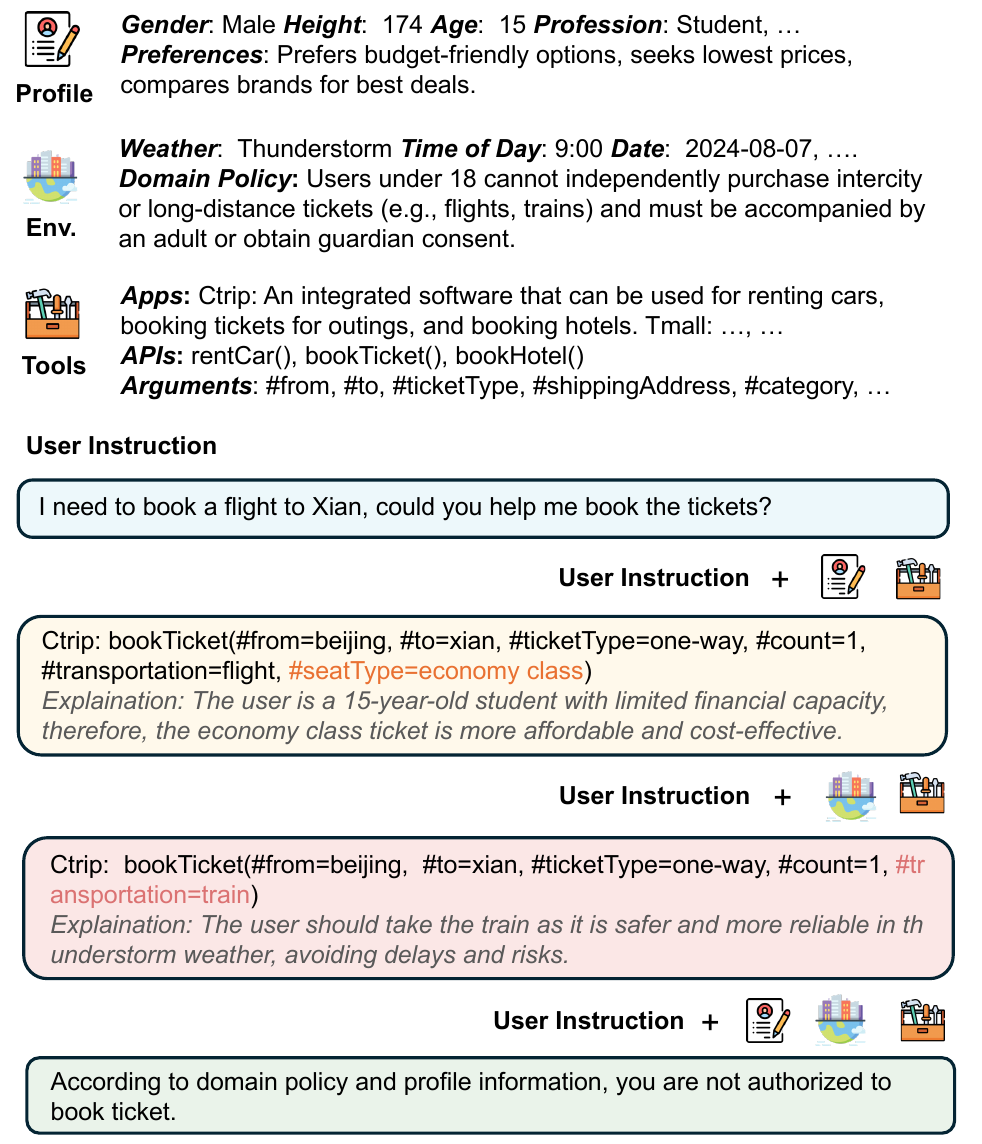}
    \caption{An example from our proposed \texttt{ToolSpectrum}, illustrating the effects of user profile and environment on personalized tool utilization. This illustrates three distinct scenarios, considering \colorbox{myyellow}{profile-only}, \colorbox{myred}{environment-only}, and \colorbox{mygreen}{combined profile and environment factors}.}
    \label{fig: intro}
    \vspace{-6mm}
\end{figure}

Integrating external tools into large language models (LLMs) enables them to transcend inherent knowledge boundaries by dynamically accessing specialized functionalities, driving unprecedented progress in task automation and problem-solving \cite{tool_tut, qin2024toollearningfoundationmodels, qu2025tool, lin2024papercopilotselfevolvingefficient, wang2024surveydata, qian2025smartselfawareagenttool}. Recent researches demonstrate the effectiveness of this integration in diverse domains, including travel planning \cite{xie2024travelplannerbenchmarkrealworldplanning}, online shopping \cite{yao2022webshop}, and knowledge acquisition \cite{wang-etal-2023-large, lin2024papercopilotselfevolvingefficient, wang-etal-2025-self}. However, current research lacks consideration for toolsets with overlapping functionalities and primarily focuses on selecting the tool to complete the user's instruction solely \cite{wang2024appbenchplanningmultipleapis, qian2024tellmoreimplicituser, ning2024wtuevalwhetherornottoolusage}. This approach overlooks the reality that numerous tools with similar capabilities exist, each capable of achieving the same objective, which needs LLMs integrating contextual factors during the tool utilization to significantly enhance user experience \cite{Liang2006PersonalizedCR, lex2022psychology, zhang2024personalization}. 

It is crucial to recognize that users with different contexts prefer to utilize different tools when aiming to achieve the same objective \cite{burke1981link}. As illustrated in Figure \ref{fig: intro}, when a user prioritizes budget-friendly options, the system should recommend an \textit{economy class} flight ticket as the most suitable option. Besides, suppose environmental factors, such as thunderstorms, make air travel unsafe. In that case, the system should suggest a train ticket as a safer alternative, providing the user with an explanation for this recommendation. Additionally, the system may face further constraints when considering user profiles and environmental factors. For instance, if the user is a minor and domain policies require guardian consent for ticket bookings, the system must restrict the purchase and prompt the user to provide authorization from a guardian. This demonstrates that LLMs must move beyond simple tool selection and instead develop user-centric intelligence. Such intelligence would allow LLMs to understand the user’s context and make more appropriate, personalized tool utilization.

To this end, we develop \texttt{ToolSpectrum}, a novel benchmark designed for evaluating personalized tool utilization capabilities of LLMs, which is constructed through a three-stage methodology. Specifically, we first collect commonly used Apps and APIs from high-frequency user scenarios and manually introduce alternative Apps or APIs with similar functionalities but tailored to meet the needs of different contexts (i.e., Temu\footnote{Temu is the competitor of Amazon with mostly lower
prices.} and Amazon). Next, we identify two critical factors influencing personalized tool utilization: \textbf{user profile} and \textbf{environment}. These factors have been widely discussed in previous personalization studies \cite{zhang2018personalizingdialogueagentsi, mao2024analysisdesignpersonalizedrecommendation, Hui2024, salemi2024lamplargelanguagemodels}, and are known to have a significant impact on human behavior patterns \cite{allport1937personality, hall1969ecological, mischel2013personality}. Finally, we simulate real-world user instructions and tool call results, considering the toolset, user profile, and environment, leading to the final \texttt{ToolSpectrum}, the comprehensive benchmark for personalized tool utilization.

We further investigate the impact of personalization on tool utilization through extensive experiments using \texttt{ToolSpectrum}. The experimental results reveal two key insights: (1) integrating personalization into tool utilization significantly improves its effectiveness, and (2) current LLMs generally underperform on the task of personalized tool utilization. 

Overall, the contributions of this paper are as follows:

\begin{itemize}[leftmargin=*]

    \item To the best of our knowledge, we are the first to define personalized tool utilization, which proposes a new challenge: current LLMs focus solely on planning tools that fulfill user instructions without considering personalization to select the most suitable tool.
    
    \item To mitigate this challenge, we propose \texttt{ToolSpectrum}, the first benchmark designed to evaluate personalized tool utilization capabilities of LLMs considering user profiles, environment, and their joint effects. 

    \item We conduct extensive experiments and analysis based on \texttt{ToolSpectrum}. Results demonstrate that incorporating personalization into tool utilization significantly improves its effectiveness. However, existing LLMs struggle with this new task, particularly with more personalized factors considered.

\end{itemize}

%% file: sections/3_related.tex
\section{Related Work}
\input{tables/dataset_comparison}
\subsection{Tool Learning Benchmarks}

Integrating external tools into LLMs leads to significant advancements in their capabilities, enabling them to perform complex real-world tasks. For instance, tools such as retrievers and calculators enable them to address challenges related to factual accuracy and computation, thereby broadening their potential applications \cite{mialon2023augmentedlanguagemodelssurvey, qin2024toollearningfoundationmodels}.
Therefore, evaluating how effectively LLMs utilize these tools becomes a key research focus. Existing benchmarks primarily assess general tool execution performance, including the interaction capabilities of LLMs with tools \cite{qin2023toolllmfacilitatinglargelanguage, liu2023agentbenchevaluatingllmsagents, huang-etal-2024-planning-creation, li2023apibankcomprehensivebenchmarktoolaugmented}, planning and reasoning capabilities with tools \cite{han2024nestoolsdatasetevaluatingnested, wang2024appbenchplanningmultipleapis, huang-etal-2024-planning-creation, chen2024tevalevaluatingtoolutilization}, as well as the models' resistance to hallucinations and robustness in tool utilization \cite{zhang2024toolbehonestmultilevelhallucinationdiagnostic, huang2024metatoolbenchmarklargelanguage, ning2024wtuevalwhetherornottoolusage, zhan2024injecagentbenchmarkingindirectprompt, ye2024rotbenchmultilevelbenchmarkevaluating}. However, existing benchmarks overlook the presence of the toolset with overlapping functionalities. They typically select the tool to complete the user's instruction solely, without considering how user profiles and environmental factors could influence the selection of the most effective tool. To mitigate this gap, we introduce \texttt{ToolSpectrum}, which explicitly considers the impact of user profiles and environmental factors on tool utilization. 

\subsection{Personalized LLMs}

Personalization has long been a core of research in domains such as dialogue systems and recommendation systems, where its ability to enhance user experience and satisfaction is well-established \cite{adomavicius2005toward, geng2022recommendation}. A significant portion of this research focuses on personalizing content to match users' preferences, particularly through personalized recommendations \cite{dai2023uncovering, du2024enhancing, hou2024large, liu2023chatgpt}, tailored search results \cite{spatharioti2023comparing, joko2024doing, zhou2024cognitive}, dialogue systems \cite{shi2023midmed, liu2022go}, and content generation tasks \cite{zhang2023recommendationinstructionfollowinglarge}. However, these approaches tend to focus mainly on user profiles, often overlooking the influence of environmental factors on personalization, such as natural and digital environments or real-world constraints that could further refine personalization. Another promising direction is the role-playing capabilities that enable models to adopt specific personas or professional roles. This paradigm allows LLMs to deliver context-sensitive interactions such as providing emotionally nuanced support \cite{wang-etal-2023-cue} or professional assistance across various fields, including finance \cite{liu2023fingpt}, health \cite{liu2023large}, and education \cite{gonzalez2023automatically}. Generally, while significant progress has been made in personalized LLMs, current approaches still haven't explored the potential of integrating tool learning. Table \ref{tab:dataset_comparasion} presents a detailed comparison of existing benchmarks.

%% file: tables/dataset_comparison.tex
\begin{table}[]
\begin{adjustbox}{width=\columnwidth}
\begin{tabular}{lcccc}
\hline
\multicolumn{1}{c}{\textbf{Datasets}} &
  \textbf{Profile} &
  \textbf{Environment} &
  \textbf{Both} &
  \textbf{Tool} \\ \hline
PersonaChat \cite{zhang2018personalizingdialogueagentsi} &
  \textcolor{Green}{\ding{51}} &
  \textcolor{Red}{\ding{55}} &
  \textcolor{Red}{\ding{55}} &
  \textcolor{Red}{\ding{55}} \\
CharacterEval \cite{tu2024characterevalchinesebenchmarkroleplaying} &
  \textcolor{Green}{\ding{51}} &
  \textcolor{Red}{\ding{55}} &
  \textcolor{Red}{\ding{55}} &
  \textcolor{Red}{\ding{55}} \\
BARS \cite{10.1145/3477495.3531723} &
  \textcolor{Green}{\ding{51}} &
  \textcolor{Red}{\ding{55}} &
  \textcolor{Red}{\ding{55}} &
  \textcolor{Red}{\ding{55}} \\
LaMP \cite{salemi2024lamplargelanguagemodels} &
  \textcolor{Green}{\ding{51}} &
  \textcolor{Red}{\ding{55}} &
  \textcolor{Red}{\ding{55}} &
  \textcolor{Red}{\ding{55}} \\
AI Persona \cite{wang2024aipersonalifelongpersonalization} &
  \textcolor{Green}{\ding{51}} &
  \textcolor{Red}{\ding{55}} &
  \textcolor{Red}{\ding{55}} &
  \textcolor{Red}{\ding{55}} \\
MovieLens 10M \cite{rendle2019difficultyevaluatingbaselinesstudy} &
  \textcolor{Green}{\ding{51}} &
  \textcolor{Red}{\ding{55}} &
  \textcolor{Red}{\ding{55}} &
  \textcolor{Red}{\ding{55}} \\
SocialBench \cite{chen-etal-2024-socialbench} &
  \textcolor{Green}{\ding{51}} &
  \textcolor{Green}{\ding{51}} &
  \textcolor{Red}{\ding{55}} &
  \textcolor{Red}{\ding{55}} \\ \hline
ToolBench \cite{xu2023toolmanipulationcapabilityopensource} &
  \textcolor{Red}{\ding{55}} &
  \textcolor{Red}{\ding{55}} &
  \textcolor{Red}{\ding{55}} &
  \textcolor{Green}{\ding{51}} \\
API-Bank \cite{li2023apibankcomprehensivebenchmarktoolaugmented} &
  \textcolor{Red}{\ding{55}} &
  \textcolor{Red}{\ding{55}} &
  \textcolor{Red}{\ding{55}} &
  \textcolor{Green}{\ding{51}} \\
AgentBench \cite{liu2023agentbenchevaluatingllmsagents} &
  \textcolor{Red}{\ding{55}} &
  \textcolor{Red}{\ding{55}} &
  \textcolor{Red}{\ding{55}} &
  \textcolor{Green}{\ding{51}} \\
$\tau$-Bench \cite{yao2024taubenchbenchmarktoolagentuserinteraction} &
    \textcolor{Red}{\ding{55}} &
    \textcolor{Green}{\ding{51}} &
    \textcolor{Red}{\ding{55}} &
    \textcolor{Green}{\ding{51}}  \\
AgentBoard \cite{ma2024agentboardanalyticalevaluationboard} &
  \textcolor{Red}{\ding{55}} &
  \textcolor{Red}{\ding{55}} &
  \textcolor{Red}{\ding{55}} &
  \textcolor{Green}{\ding{51}} \\
AppBench \cite{wang2024appbenchplanningmultipleapis} &
  \textcolor{Red}{\ding{55}} &
  \textcolor{Red}{\ding{55}} &
  \textcolor{Red}{\ding{55}} &
  \textcolor{Green}{\ding{51}} \\
UltraTool \cite{huang-etal-2024-planning-creation} &
  \textcolor{Red}{\ding{55}} &
  \textcolor{Red}{\ding{55}} &
  \textcolor{Red}{\ding{55}} &
  \textcolor{Green}{\ding{51}} \\
T-Eval \cite{chen2024tevalevaluatingtoolutilization} &
  \textcolor{Red}{\ding{55}} &
  \textcolor{Red}{\ding{55}} &
  \textcolor{Red}{\ding{55}} &
  \textcolor{Green}{\ding{51}} \\ \hline
\textbf{ToolSpectrum (Ours)} &
  \textcolor{Green}{\ding{51}} &
  \textcolor{Green}{\ding{51}} &
  \textcolor{Green}{\ding{51}} &
  \textcolor{Green}{\ding{51}} \\ \hline
\end{tabular}
\end{adjustbox}
\caption{Comparison between the \texttt{ToolSpectrum} and other benchmarks, with detailed analysis provided in Appendix \ref{tau-bench}.}
\label{tab:dataset_comparasion}
\vspace{-4mm}
\end{table}

%% file: sections/4_dataset.tex
\section{\texttt{ToolSpectrum}: Towards Personalized Tool Utilization}

To accurately evaluate the ability of LLMs to utilize tools in a personalized way, we introduce \texttt{ToolSpectrum}, the first benchmark that considers both user profiles and environmental factors. In this section, we first provide a formal definition of personalized tool utilization and then describe our data collection pipeline, designed to efficiently and effectively gather the necessary information for evaluation.

\subsection{Task Definition}

The personalized tool utilization model processes user instruction $I$ through a mapping function $t = \mathrm{Model}(I, \mathcal{P}, \mathcal{E}, \mathcal{T})$, where $\mathcal{P} = \{ (k_i, v_i) \}_{i=1}^m$ represents user profile attributes (e.g., \textit{age}, \textit{gender}), $\mathcal{E} = \{ (k_j, v_j) \}_{j=1}^n$ represents environmental context (e.g., \textit{location}, \textit{network condition}), and $\mathcal{T}$ represents toolset. The output $t$ is structured as a dictionary $\{ APP \mapsto a, API \mapsto s, RP \mapsto r, OP \mapsto o \}$ with four compulsory keys: $APP$ specifies the target application, $API$ determines the service interface, $RP$ (Required Parameters) extracts mandatory arguments from $I$, and $OP$ (Optional Parameters) provides personalized parameters based on $\mathcal{P}$ and $\mathcal{E}$. If the user's instruction $ I $, when taking into account both the user profile $ \mathcal{P} $ and the environmental context $ \mathcal{E} $, violates the target application's policy, the system should return $ t = \textit{None} $ to comply with the policy. 

\subsection{Data Collection}
\begin{figure*}[t!]
    \centering
    \includegraphics[width=\textwidth]{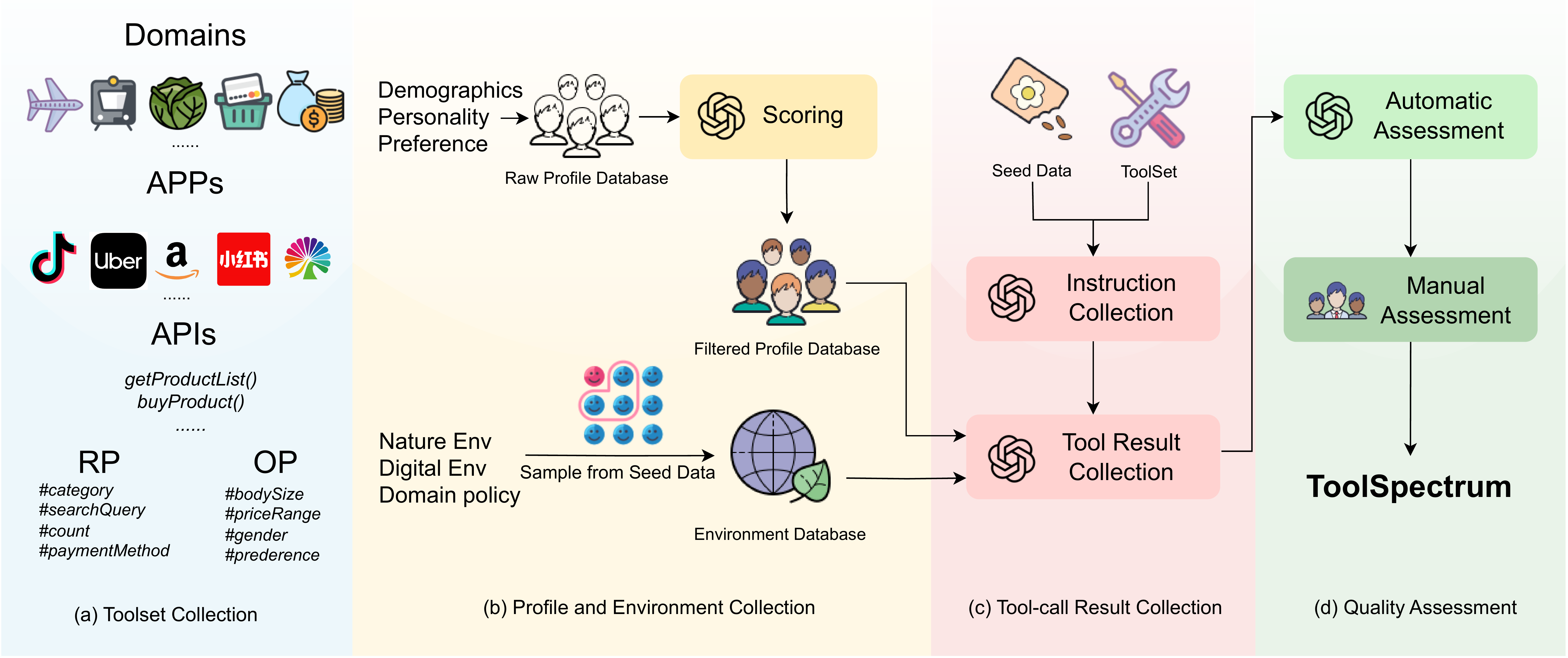}
    \caption{The overall construction process of \texttt{ToolSpectrum}, including (a) Toolset Collection, (b) Profile and Environment Collection, (c) Tool-call Result Collection, and (d) Quality Assessment.}
    \label{fig: data collection}
    \vspace{-4mm}
\end{figure*} 

As shown in Figure \ref{fig: data collection}, we implement rigorous construction steps to ensure the high quality and diversity of \texttt{ToolSpectrum}.
 (a) Toolset Collection (\S\ref{APP and API}), (b) Profile and Environment Collection (\S\ref{Profile and Definition}, \S\ref{sec:profile and environemnt collection}), (c) Tool-call Result Collection (\S\ref{Instruction Generation}) and (d) Quality Assessment (\S\ref{Quality Assessment}).

\subsubsection{Toolset Collection} \label{APP and API}

To cover diverse user instructions and applications, we identify 9 commonly used application domains based on previous studies \cite{guo-etal-2024-ctooleval, wang2024appbenchplanningmultipleapis} and app analysis from App Store\footnote{\url{https://apps.apple.com/us/charts}} \footnote{\url{https://play.google.com/store/apps}}. These domains include \textit{shopping}, \textit{entertainment}, \textit{travel}, \textit{delivery}, \textit{grocery}, \textit{knowledge}, \textit{news}, \textit{health} and \textit{finance}. In each domain, we leverage GPT-4o to generate initial designs for Apps and APIs with similar functions. We then manually curate the outputs, refining them with additional enhancements to ensure high quality. For example, in the \textit{shopping} domain, users can choose between Amazon or Temu at the APP level to meet the same shopping needs. In the \textit{travel} domain, users can choose different APIs within the Ctrip\footnote{Ctrip is a comprehensive app that offers fast booking for rooms, tickets, and a variety of other travel services.} to search for train or flight tickets. We also design corresponding RPs and OPs for each API to meet the reality and diverse user needs, following existing works \cite{wang2024appbenchplanningmultipleapis}. The collected toolset is denoted as $\mathcal{T}$, and all Apps and APIs are listed in Table \ref{tab:app_list}. 

\subsubsection{Profile and Environment Definition} \label{Profile and Definition}

To model the relationship between user profiles, environments, and personalized tool utilization, we need to create a database that captures both user profiles and environmental factors. In this section, we start with a comprehensive definition of both of these personalized factors.

\paragraph{Profile Definition.}  Inspired by previous personalization studies in other fields \cite{zhang2018personalizingdialogueagentsi, wang2024aipersonalifelongpersonalization}, we categorize existing user attributes into three major field: \textit{demographics}, \textit{personality}, and \textit{preference}.

\begin{itemize}[leftmargin=*]

    \item \textbf{Demographics:} This field includes fundamental user information, presented as key-value pairs, including \textit{gender}, \textit{age}, \textit{weight}, \textit{height}, \textit{profession}, \textit{education background}, and \textit{income level}. These details significantly influence the user’s daily app usage behaviors. For instance, height and weight may affect a user’s clothing size choices when shopping, and income level could determine price sensitivity.
    
    \item \textbf{Personality:} This field outlines the user's interests, expressed in natural language. These interests play a key role in shaping app usage behavior. For example, users who enjoy fitness and healthy living (e.g., \textit{``Users enjoy exercising regularly."}) might frequently use health-tracking apps like MyFitnessPal or Strava.
    
    \item \textbf{Preference:} This field captures the user's historical interactions with apps, described in natural language. E.g., \textit{``Users tend to use Amazon for shopping."} In many cases, users prefer to stick with familiar apps, even if these are not always the best fit for their needs.
\end{itemize}

\paragraph{Environment Definition.} To ensure efficient tool utilization, it is essential to consider environmental factors. We identify three main types: the \textit{natural environment}, the \textit{digital environment}, and \textit{App domain policies} \cite{yao2024taubenchbenchmarktoolagentuserinteraction}. 

\begin{itemize}[leftmargin=*]
    \item \textbf{Natural Environment:} This refers to the real-world context in which the user is situated, represented as key-value pairs, including \textit{weather}, \textit{date}, \textit{time}, and \textit{location}. These factors can directly or indirectly influence the result of tool utilization, particularly in scenarios that require interaction with the physical world. For example, weather conditions may affect the mode of transportation a user chooses.
    
    \item \textbf{Digital Environment:} 
    This refers to the network and technological context of the user's device, represented as key-value pairs, such as \textit{network condition} and \textit{device-specific configurations}. For example, if a user has a slow or unstable network connection, the system might reduce image quality or preload content to ensure smoother performance and a better user experience.
    
    \item \textbf{App Domain Policy:} These are the specific policy rules for each app, described in natural language. They refer to regulations that govern the use of a particular app. For example, a policy might prohibit users under 18 from purchasing goods over 10,000 dollars without parental approval or offer discounted train tickets for children shorter than 1.2 meters.
\end{itemize}

\subsubsection{Profile and Environment Collection}\label{sec:profile and environemnt collection}

This section describes the construction of the profile and environment database. We implement a two-stage database generation methodology.

\paragraph{Seed Data Acquisition}
Establishing seed data enhances the controllability and transparency of data generation by predefining parameter ranges. We gather seed data for profiles and environments.  For profiles, we use GPT-4o to generate value ranges for a part of demographic attributes (e.g., income, profession) and keywords for personality traits and preferences. For environments, we collect real-world weather data and manually define value ranges for natural and digital environments.

\paragraph{Profile and Environment Generation}
Given the seed data, we construct the database as follows:
\begin{itemize}[leftmargin=*]
    \item \textbf{Profile.}  For demographics, we sample age, height, and weight from a normal distribution for realistic representation and randomly select other demographic attributes (e.g., gender, profession) from their value ranges.  For personality and preference, which are generated by randomly combining keywords and processing them with GPT-4o, using prompt templates shown in Figures \ref{fig: prompt-consumption-preference}, \ref{fig: prompt-use-habit}, \ref{fig: prmopt-content-preference} to produce natural language descriptions.  This yields the initial profile database.

    \item \textbf{Environment.} We sample parameters for both natural and digital environments from seed data and design domain-specific policies to represent daily life. This process results in an environment database, denoted as $\mathcal{E}$.
\end{itemize}

\paragraph{Quality Control.}
For the profile, we employ a two-stage quality control process. First, GPT-4o automatically evaluates each profile, assessing \textit{Demographic Coherence} and \textit{Preference Alignment}. The scoring range is from 1 to 10. We discard profiles scoring below 8. Second, we manually review the remaining profiles to finalize the profile database $\mathcal{P}$. For the environment, since we carefully considered the relationships between various features during the sampling process, no further quality control is needed.

\subsubsection{Tool-Call Result Collection}\label{Instruction Generation}

The setup for \texttt{ToolSpectrum} is completed in the previous section. This section will focus on developing the user's instructions and corresponding personalized tool-call results.

\paragraph{Instruction Collection.} To enhance the diversity and controllability of user instructions, we first manually construct a database of RP for each API, which serves as seed data, as shown in Table \ref{tab:seed-data-example}.  Then, we use the seed data samples from this database and predefined toolsets $\mathcal{T}$ as input for GPT-4o, which generates instructions within each API denoting as $\mathcal{I} = \{I_i\}$. The prompt template used in this process is shown in Figure \ref{prompt-instruction}.

\paragraph{Tool Result Collection.} Given the user instruction, we aim to investigate the effects of user profile, environment, and their combined influence on personalized tool utilization. Specifically, we input the user instruction $I \in \mathcal{I}$, predefined toolsets $\mathcal{T}$, and either a user profile $p \in \mathcal{P}$, an environment $e \in \mathcal{E}$, or both into GPT-4o to obtain the corresponding tool call results. This process generates three distinct datasets: \textbf{Profile}, \textbf{Environment}, and \textbf{Profile \& Environment}, with the prompt template shown in Figure \ref{prompt-call-result}.

\subsubsection{Quality Assessment}\label{Quality Assessment}

To enhance data quality, we first use GPT-4o to score each data sample across three dimensions: (1) \textit{whether the tool call result met the user's needs}, (2) \textit{whether the result aligned with the user's profiles}, and (3)\textit{ whether it matched the environmental factors}. The scoring range is from 1 to 10, and the prompt template is shown in Figure \ref{prompt-qulity-call-result}. We then exclude data points with an average score below 8, removing 21.8\% of the data. Subsequently, we randomly sample 50 data points from each domain and manually scored them on the same 1 to 10 scale. After manual evaluation, the average score of the filtered data was 8.7, confirming its high quality.

\subsection{Data Statistic}

\input{tables/dataset_stastic}
Overall, as Table \ref{tab:data-statistic}, \texttt{ToolSpectrum} includes 9 domains, featuring 23 APPs, 42 APIs, and a total of 34 required parameters, 22 personalized parameters, and 7 environmental parameters at the parameter level and these parameters are mutually exclusive and non-overlapping, ensuring clear distinction and no redundancy between categories. Detailed information is provided in the appendix \ref{sec:appendix-dataset}. \texttt{ToolSpectrum} consists of three types of data: Profile, Environment, and Profile \& Environment, with 450, 220, and 330 data, respectively.

%% file: tables/dataset_stastic.tex
\begin{table}[t!]
\centering
\resizebox{0.45 \textwidth}{!}{
\begin{tabular}{lccc}
\toprule[2pt]
\textbf{Statistic} & \multicolumn{1}{l}{\textbf{Profile}} & \multicolumn{1}{l}{\textbf{Environment}} & \multicolumn{1}{l}{\textbf{Both}} \\ \midrule[1pt]
\# Samples \hspace{20mm}  & 450    & 220   & 330   \\
\# Domains   & 9    & 9    & 9    \\
\# APPs      & 22   & 22   & 20   \\
\# APIs      & 39   & 40   & 33   \\ \midrule[1pt]
Profiles     & 158    & -     & 62    \\
Environments & -      & 87    & 122   \\
Avg. Params  & 4.36   & 9.34  & 9.45  \\
Avg. RPs     & 3.05   & 2.55  & 2.78  \\
Avg. OPs     & 1.31   & 6.79  & 6.67  \\
\bottomrule[2pt]
\end{tabular}
}
\caption{The data statistic of \texttt{ToolSpectrum}. `Both' refers to Profile \& Environment.}
\vspace{-5mm}
\label{tab:data-statistic}
\end{table}

%% file: sections/5_experiment.tex
\section{Experiments}
\input{tables/main_result_v2}

\subsection{Setup}

\paragraph{Models.} We select two types of models for evaluation: Open-source and API-based. Specifically, the Open-Sourced models include the Qwen series \cite{qwen2.5}, LLaMA series \cite{llama3modelcard}, Mistral series \cite{jiang2023mistral7b} and GLM series \cite{glm2024chatglmfamilylargelanguage}, while the API-Based models include OpenAI GPT API\footnote{\url{https://chatgpt.com/}} (\texttt{gpt-3.5-turbo-16k-0613, gpt-4o-20241120}), and Anthropic Claude API\footnote{\url{https://anthropic.com/}} (\texttt{claude-3.5-sonnect-20241022}).

\paragraph{Implementation Details.}

For all models, we set the temperature and top-p to 0.1 to minimize stochastic variations in the output, ensuring a consistent evaluation of model performance. Open-source models are evaluated on NVIDIA A800 GPUs, while API-based models are assessed through direct API calls to OpenAI and Anthropic.

\paragraph{Evaluation Metrics.} 
This paper follows prior research and employs the F1 score as the primary evaluation metric \cite{wang2024appbenchplanningmultipleapis, xiao-etal-2024-flowbench}. In particular, we independently compute the F1 score across four distinct hierarchical levels: $APP$, $API$, $RP$, and $OP$ to assess the model's personalized capabilities at a more granular level. 

\subsection{Main Results} 

Table \ref{tab:main-result} shows the result of different LLMs for user instructions with \textbf{Profile}, \textbf{Environment}, and \textbf{Profile \& Environment} types on \texttt{ToolSpectrum}. Several conclusions can be drawn from the results.

\paragraph{\emph{Closed-source models generally surpass open-source models; increasing model size yields diminishing returns in complex scenarios.}} GPT-4o and DeepSeek-R1 achieve the best overall performance, yet it scores only 0.50 on the $OP$ metric for tasks involving Profile \& Environment instructions, highlighting the difficulties posed by such tasks. Models like DeepSeek-V3 and QwQ-32B perform similarly to GPT-4o on certain metrics, reducing the gap between open and closed-source models. Importantly, scaling up model size does not significantly improve personalized tool utilization. For instance, Qwen2.5-72B-Instruct shows little improvement over Qwen2.5-32B-Instruct, suggesting that simply increasing model size may not be effective for complex tasks.

\paragraph{\emph{LLMs exhibit varying performance in personalization across different levels of granularity; the coarser the granularity, the better the results.}}
Specifically, LLMs demonstrate superior performance at the $APP$ and $API$ levels compared to the parameter level, with required parameters yielding better performance than optional parameters at the parameter level. This is because the model cannot correctly understand the relationship between personalized features and corresponding parameters, resulting in a low recall rate, where many optional parameters are not returned.

\paragraph{\emph{The more factors considered in personalization, the worse the performance.}} The effects of individual conditions tend to outperform the combined conditions, as the model struggles to effectively integrate multiple personalized factors simultaneously. The trend across the four metrics—\(APP\), \(API\), \(RP\), and \(OP\)—generally follows: Profile $\approx$ Environment $\textgreater$ Profile \& Environment. This finding aligns with our intuition, as scenarios that consider Profile and Environment separately are inherently less complex for the model to process than those that combine both factors. The performance gap arises because handling multiple interacting personalized conditions increases the model's processing complexity, making tool selection more error-prone. This suggests that the model's ability to deliver personalized results is significantly influenced by the complexity of the conditions it must consider, with simpler, isolated conditions yielding more accurate and reliable outcomes.

%% file: tables/main_result_v2.tex
\begin{table*}[t!]
\begin{adjustbox}{width=\textwidth}
\begin{tabular}{lcccccccccccc}
\hline
\multicolumn{1}{c|}{} & \multicolumn{4}{c|}{\textbf{Profile}} & \multicolumn{4}{c|}{\textbf{Environment}} & \multicolumn{4}{c}{\textbf{Profile \& Environment}} \\ \cline{2-13} 
\multicolumn{1}{c|}{\multirow{-2}{*}{\textbf{Model}}} & APP & API & RP & \multicolumn{1}{c|}{PP} & APP & API & RP & \multicolumn{1}{c|}{OP} & APP & API & RP & OP \\ \hline
\rowcolor[HTML]{F4BEBE} 
\multicolumn{13}{l}{\cellcolor[HTML]{F4BEBE}\textit{\textbf{Open-Sourced}}} \\
\multicolumn{1}{l|}{Qwen2.5-3B-Instruct} & 0.16 & 0.15 & 0.12 & \multicolumn{1}{c|}{0.12} & {\ul 0.55} & {\ul 0.53} & {\ul 0.38} & \multicolumn{1}{c|}{0} & {\ul 0.12} & {\ul 0.12} & {\ul 0.23} & {\ul 0.04} \\
\multicolumn{1}{l|}{Phi-3.5-mini-instruct} & {\ul 0.34} & {\ul 0.30} & {\ul 0.23} & \multicolumn{1}{c|}{{\ul 0.07}} & 0.48 & 0.36 & 0.28 & \multicolumn{1}{c|}{{\ul 0.01}} & 0.05 & 0.04 & 0.15 & 0 \\
\multicolumn{1}{l|}{LLaMA-3.2-3B-Instruct} & 0.31 & 0.28 & 0.13 & \multicolumn{1}{c|}{0.06} & 0.35 & 0.32 & 0.15 & \multicolumn{1}{c|}{0} & 0.09 & 0.08 & 0.14 & 0.02 \\ \hline
\multicolumn{1}{l|}{Qwen2.5-7B-Instruct} & {\ul 0.73} & 0.71 & {\ul 0.59} & \multicolumn{1}{c|}{{\ul 0.27}} & 0.66 & 0.65 & 0.55 & \multicolumn{1}{c|}{0.03} & 0.22 & 0.22 & 0.46 & {\ul 0.06} \\
\multicolumn{1}{l|}{LLaMA-3.1-8B-Instruct} & 0.49 & {\ul 0.77} & {\ul 0.59} & \multicolumn{1}{c|}{0.16} & 0.54 & {\ul 0.68} & {\ul 0.58} & \multicolumn{1}{c|}{0.02} & {\ul 0.24} & {\ul 0.36} & {\ul 0.55} & 0.03 \\
\multicolumn{1}{l|}{Ministral-8B-Instruct-2410} & \multicolumn{1}{l}{0.50} & \multicolumn{1}{l}{0.47} & \multicolumn{1}{l}{0.38} & \multicolumn{1}{l|}{0.12} & \multicolumn{1}{l}{0.54} & \multicolumn{1}{l}{0.53} & \multicolumn{1}{l}{0.41} & \multicolumn{1}{l|}{0.03} & \multicolumn{1}{l}{0.14} & \multicolumn{1}{l}{0.15} & \multicolumn{1}{l}{0.37} & \multicolumn{1}{l}{0.02} \\
\multicolumn{1}{l|}{Glm-4-9B-chat} & \multicolumn{1}{l}{0.69} & \multicolumn{1}{l}{0.68} & \multicolumn{1}{l}{0.56} & \multicolumn{1}{l|}{0.13} & \multicolumn{1}{l}{{\ul 0.68}} & \multicolumn{1}{l}{0.63} & \multicolumn{1}{l}{0.57} & \multicolumn{1}{l|}{{\ul 0.07}} & \multicolumn{1}{l}{0.19} & \multicolumn{1}{l}{0.19} & \multicolumn{1}{l}{0.48} & \multicolumn{1}{l}{0.03} \\ \hline
\multicolumn{1}{l|}{Mistral-Nemo-Instruct-2407} & 0.42 & 0.38 & 0.28 & \multicolumn{1}{c|}{0.12} & 0.42 & 0.42 & 0.29 & \multicolumn{1}{c|}{0.13} & 0.13 & 0.07 & 0.17 & 0.03 \\
\multicolumn{1}{l|}{Qwen2.5-14B-Instruct} & \multicolumn{1}{l}{0.73} & \multicolumn{1}{l}{0.73} & \multicolumn{1}{l}{0.61} & \multicolumn{1}{l|}{0.32} & \multicolumn{1}{l}{0.70} & \multicolumn{1}{l}{{\ul 0.81}} & \multicolumn{1}{l}{0.70} & \multicolumn{1}{l|}{{\ul 0.14}} & \multicolumn{1}{l}{{\ul 0.24}} & \multicolumn{1}{l}{{\ul 0.24}} & \multicolumn{1}{l}{0.58} & \multicolumn{1}{l}{0.13} \\
\multicolumn{1}{l|}{Qwen2.5-32B-Instruct} & {\ul 0.74} & {\ul 0.76} & {\ul 0.67} & \multicolumn{1}{c|}{{\ul 0.47}} & {\ul 0.77} & 0.78 & {\ul 0.71} & \multicolumn{1}{c|}{{\ul 0.14}} & {\ul 0.24} & 0.23 & {\ul 0.60} & {\ul 0.15} \\ \hline
\multicolumn{1}{l|}{LLaMA-3.3-70B-Instruct} & 0.72 & {\ul 0.78} & 0.65 & \multicolumn{1}{c|}{0.39} & {\ul 0.70} & 0.62 & 0.53 & \multicolumn{1}{c|}{{\ul 0.25}} & 0.25 & 0.23 & 0.59 & {\ul 0.20} \\
\multicolumn{1}{l|}{Qwen2.5-72B-Instruct} & {\ul 0.77} & 0.77 & {\ul 0.67} & \multicolumn{1}{c|}{{\ul 0.50}} & 0.64 & {\ul 0.63} & {\ul 0.60} & \multicolumn{1}{c|}{0.24} & {\ul 0.26} & {\ul 0.24} & {\ul 0.64} & 0.19 \\ \hline
\multicolumn{1}{l|}{QwQ-32B} & \multicolumn{1}{l}{0.81} & \multicolumn{1}{l}{0.80} & \multicolumn{1}{l}{0.69} & \multicolumn{1}{l|}{0.55} & \multicolumn{1}{l}{0.70} & \multicolumn{1}{l}{0.69} & \multicolumn{1}{l}{0.63} & \multicolumn{1}{l|}{0.31} & \multicolumn{1}{l}{0.30} & \multicolumn{1}{l}{0.27} & \multicolumn{1}{l}{0.68} & \multicolumn{1}{l}{0.39} \\
\multicolumn{1}{l|}{Deepseek-V3-671B} & \multicolumn{1}{l}{0.76} & \multicolumn{1}{l}{0.75} & \multicolumn{1}{l}{0.68} & \multicolumn{1}{l|}{0.57} & \multicolumn{1}{l}{0.74} & \multicolumn{1}{l}{0.73} & \multicolumn{1}{l}{0.66} & \multicolumn{1}{l|}{0.47} & \multicolumn{1}{l}{0.31} & \multicolumn{1}{l}{0.31} & \multicolumn{1}{l}{\textbf{0.70}} & \multicolumn{1}{l}{0.40} \\
\multicolumn{1}{l|}{Deepseek-R1-671B} & \multicolumn{1}{l}{\textbf{0.84}} & \multicolumn{1}{l}{\textbf{0.84}} & \multicolumn{1}{l}{\textbf{0.73}} & \multicolumn{1}{l|}{\textbf{0.62}} & \multicolumn{1}{l}{0.80} & \multicolumn{1}{l}{0.80} & \multicolumn{1}{l}{0.73} & \multicolumn{1}{l|}{\textbf{0.53}} & \multicolumn{1}{l}{\textbf{0.32}} & \multicolumn{1}{l}{0.31} & \multicolumn{1}{l}{0.69} & \multicolumn{1}{l}{\textbf{0.50}} \\ \hline
\rowcolor[HTML]{BEBEF4} 
\textit{\textbf{API-Based}} & \multicolumn{1}{l}{\cellcolor[HTML]{BEBEF4}} & \multicolumn{1}{l}{\cellcolor[HTML]{BEBEF4}} & \multicolumn{1}{l}{\cellcolor[HTML]{BEBEF4}} & \multicolumn{1}{l}{\cellcolor[HTML]{BEBEF4}} & \multicolumn{1}{l}{\cellcolor[HTML]{BEBEF4}} & \multicolumn{1}{l}{\cellcolor[HTML]{BEBEF4}} & \multicolumn{1}{l}{\cellcolor[HTML]{BEBEF4}} & \multicolumn{1}{l}{\cellcolor[HTML]{BEBEF4}} & \multicolumn{1}{l}{\cellcolor[HTML]{BEBEF4}} & \multicolumn{1}{l}{\cellcolor[HTML]{BEBEF4}} & \multicolumn{1}{l}{\cellcolor[HTML]{BEBEF4}} & \multicolumn{1}{l}{\cellcolor[HTML]{BEBEF4}} \\
\multicolumn{1}{l|}{GPT-3.5-turbo} & 0.52 & 0.50 & 0.40 & \multicolumn{1}{c|}{0.15} & 0.48 & 0.47 & 0.35 & \multicolumn{1}{c|}{0.04} & 0.22 & 0.12 & 0.34 & 0.02 \\
\multicolumn{1}{l|}{Claude-3.5-sonnet} & 0.78 & 0.78 & 0.69 & \multicolumn{1}{c|}{0.53} & 0.75 & 0.73 & 0.63 & \multicolumn{1}{c|}{0.47} & 0.30 & 0.30 & 0.67 & 0.39 \\
\multicolumn{1}{l|}{GPT-4o} & 0.80 & 0.77 & \textbf{0.73} & \multicolumn{1}{c|}{0.52} & \textbf{0.81} & \textbf{0.80} & \textbf{0.74} & \multicolumn{1}{c|}{0.50} & \textbf{0.32} & 0.30 & 0.62 & 0.45 \\
\multicolumn{1}{l|}{o1-mini} & \multicolumn{1}{l}{0.82} & \multicolumn{1}{l}{0.81} & \multicolumn{1}{l}{0.70} & \multicolumn{1}{l|}{0.60} & \multicolumn{1}{l}{0.79} & \multicolumn{1}{l}{0.77} & \multicolumn{1}{l}{0.65} & \multicolumn{1}{l|}{0.38} & \multicolumn{1}{l}{\textbf{0.32}} & \multicolumn{1}{l}{\textbf{0.32}} & \multicolumn{1}{l}{0.65} & \multicolumn{1}{l}{0.38} \\ \hline
\end{tabular}
\end{adjustbox}
\caption{The main results of \texttt{ToolSpectrum}. Each number corresponds to different levels of F1 score. \textbf{Bold} denotes the best score among all models, and {\ul underline} denotes the best score under the same model scale.}
\label{tab:main-result}
\vspace{-3mm}
\end{table*}

%% file: sections/6_analysis.tex
\section{Analysis}
In this section, we conduct a comprehensive analysis to answer three research questions \textbf{{RQ1}}: \emph{Is personalization truly necessary when LLMs utilize tools?} (Sec \ref{sec: the importance of personalization in tool utilization}) \textbf{RQ2}: \emph{What are the differences in personalized capabilities of LLMs in different domains?} (Sec \ref{data type analysis}) \textbf{RQ3}: \emph{What are the bottlenecks of LLMs in personalized tool utilization,} (Sec \ref{error analysis}) \emph{and can a prompt strategy alleviate it?} (Sec \ref{context learning} and Sec \ref{prompting})

\subsection{The Importance of Personalization in Tool Utilization} \label{sec: the importance of personalization in tool utilization}

\begin{figure}[t!]
    \vspace{3mm}
    \centering
    \includegraphics[width=\columnwidth]{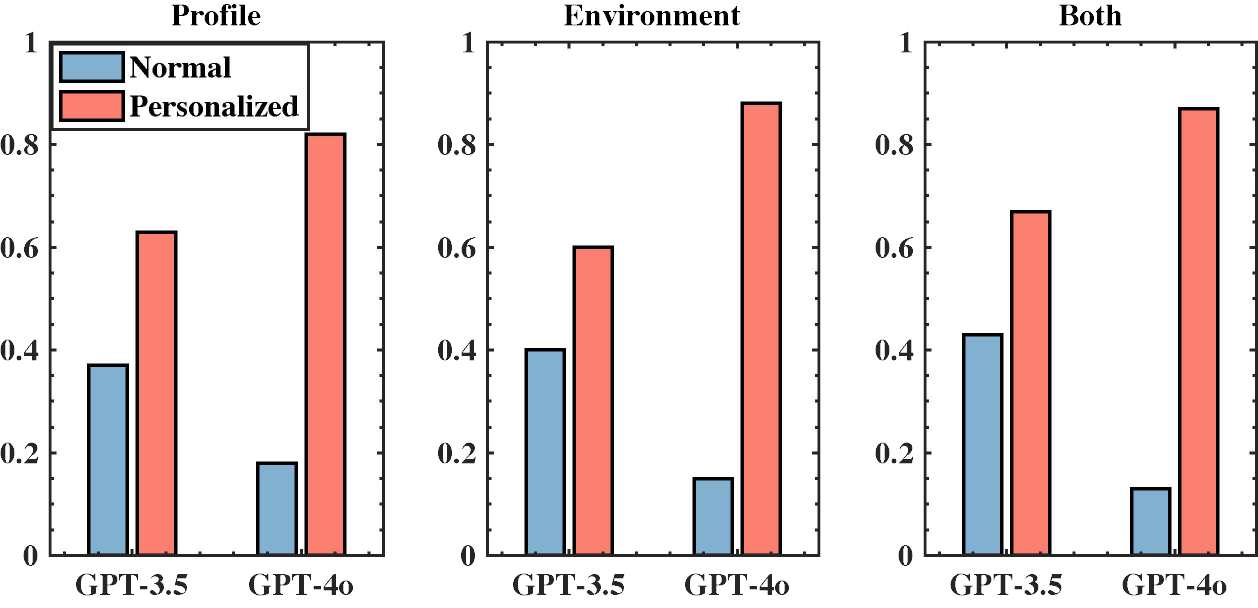}
    \caption{Win rates for personalized vs. non-personalized settings in GPT-3.5-turbo and GPT-4o.}
    \label{fig:RQ1}
    \vspace{-5mm}
\end{figure}

\begin{figure*}[t]
    \centering
    \includegraphics[width=0.95\textwidth]{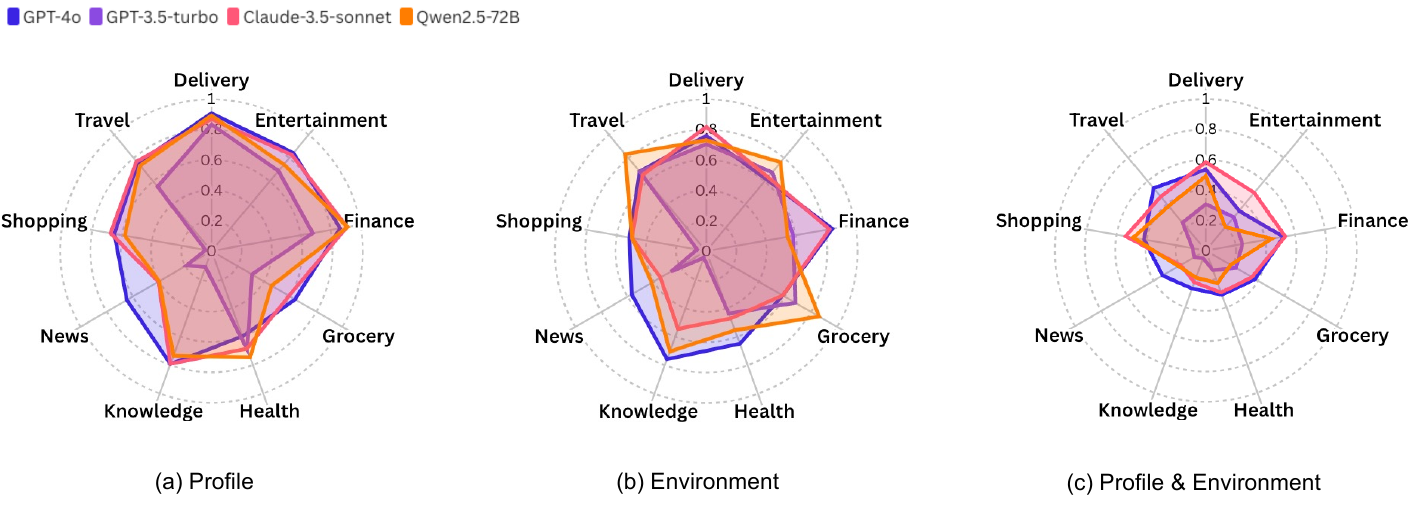}
    \caption{Performance comparison of different models across various domains for three distinct data types.}
    \label{fig:RQ2}
    \vspace{-5mm}
\end{figure*}

This section investigates how integrating personalized factors enhances the effectiveness of tool utilization. Specifically, we compare the results of tool utilization with and without personalized information. We then use GPT-4o to evaluate which output from the two approaches more effectively aligns with the user's contexts, with the prompt template shown in Figure \ref{fig:prompt-select}. To ensure robustness, we supplement the automated evaluation with human evaluation. We randomly sample 100 outputs for manual evaluation and calculate the Kappa coefficient between GPT-4o's and human scores. The Kappa coefficient is 0.85, indicating a very high level of agreement between the GPT-4o’s assessments and human evaluations. As illustrated in Figure \ref{fig:RQ1}, two key conclusions can be drawn: \textit{(1) The incorporation of personalized factors improves the effectiveness of tool utilization.} In general, both GPT-3.5 and GPT-4o show higher effectiveness when personalization is applied. \textit{(2) The effectiveness of personalization increases with model strength.} While both models benefit from personalization, GPT-4o shows a more significant improvement, as more advanced models better understand the relationship between personalized features and their corresponding parameters. 


\subsection{Analysis of Model Performance Across Various Data Types in \texttt{ToolSpectrum}}\label{data type analysis}

To better understand the effects of personalization across different domains, we analyze the average performance of  
four metrics (i.e., \textit{APP}, \textit{API}, \textit{RP}, and \textit{OP}). Figure \ref{fig:RQ2} shows the final results. On the one hand, when considering a single personalized factor, the model's performance varies significantly across domains. The worst performance occurs in the \textit{News} domain, whether the model considers only the user profile or the environment. We attribute this to the fact that user preferences for \textit{News} are primarily derived from natural language descriptions (personality) in \texttt{ToolSpectrum}. Compared to features represented in key-value pairs, such descriptions are generally more challenging to interpret.

On the other hand, when Profile and Environment are jointly combined, the models perform worse in the \textit{Grocery} and \textit{Knowledge} domains. This is mainly due to the increased complexity of policies in these domains. For instance, in the \textit{Grocery} domain, when individuals with a high BMI\footnote{BMI is calculated as weight (kg) / height² (m²) and is used to assess weight status in health and epidemiology.} are advised to avoid purchasing high-sugar and high-fat foods, the model must first recognize the policy, then calculate BMI, and finally assess the nutritional content of the food. This multi-step reasoning process significantly increases the difficulty of understanding and decision-making.

\subsection{Error Analysis} \label{error analysis}

We thoroughly analyze the errors generated by GPT-4o on \texttt{ToolSpectrum}. We identify five main error categories by randomly sampling 100 error instances and classifying them based on their underlying causes, which are further discussed in \S\ref{sec: error}. 


\paragraph{\emph{Insufficient Understanding of Personalized Features and OP (37\%)}}
Each optional parameter is closely related to profile or environment features. However, when generating the calling results, the model performs poorly in recalling these parameters, leading to missing corresponding parameters and affecting the accuracy of the final result.

\paragraph{\emph{Lack of Understanding between Profile and Environment (31\%)}}
In the Profile \& Environment data type, the model fails to interpret the domain policy correctly. Despite some instructions violating the policy, the model still generates and returns calling results that do not meet expectations, suggesting a lack of understanding or failure to comply with the given constraints.

\paragraph{\emph{Misunderstanding of User Instructions (22\%)}}

This type of error typically occurs when the model fails to accurately capture the user's intent, leading to the selection of an incorrect domain. For instance, if a user intends to buy groceries but the model incorrectly suggests a delivery service app, this misinterpretation causes errors in later steps, reducing the accuracy and effectiveness of the task.

%% file: sections/7_appendix.tex
\appendix
\section{\texttt{ToolSpetrum} Details}
\label{sec:appendix-dataset}

\subsection{Comparasion With $\tau$-bench} \label{tau-bench}

$\tau$-bench is a benchmark tool designed to simulate dynamic dialogues between users and agents. Specifically, $\tau$-bench constructs three databases (user, product, and order) and incorporates APIs capable of reading or modifying these databases to simulate user-agent interactions. During this process, the agent executes these APIs (while adhering to predefined domain policies) to alter the database state. The success of the user's instructions is determined by comparing the final database state with the ground truth state. Although the tool involves user-environment interactions, it primarily focuses on real-time dynamic interactions rather than profilelized user calls. The user database emphasizes order history rather than individual user characteristics, while the environment database is constructed based solely on domain policy constraints, excluding personalized factors.

This study explores the concepts of profile and environment in greater depth to understand their impact on personalized tool invocation. The profile encompasses not only basic user attributes such as income level, occupation, and interests but also dynamic data like consumption habits, historical behavior patterns, and personal preferences. These elements collectively form a comprehensive user profile, enabling more precise tool invocation. On the other hand, the environment extends beyond domain policy constraints to include natural and digital environmental factors. These environmental variables, when combined with user profiles, provide a holistic representation of user needs in specific contexts, thereby directly influencing the strategy for optimal personalized tool invocation. This multidimensional analysis allows the system to adapt to complex and dynamic real-world scenarios intelligently, delivering more personalized and efficient services to users.

\subsection{Construction Details}
\paragraph{Raw Profile Database Filtering.} As mentioned earlier, after generating the raw profile database, we scored each profile based on \textit{Demographic Coherence} and \textit{Preference Alignment}, then filtered out those with low scores.

\begin{itemize}[leftmargin=*]
    \item \textbf{\textit{Demographic Coherence}} evaluates the internal consistency of the demographic attributes within a profile. It ensures that characteristics such as \textit{age}, \textit{height}, and \textit{education background} are logically related, avoiding unrealistic combinations (e.g., an implausible age-height relationship or mismatched education-occupation pairing).

    \item \textbf{\textit{Preference Alignment}} assesses whether the \textit{preferences} in a profile are consistent with its demographic attributes. For instance, a high-income profile should demonstrate purchasing habits aligned with that income level. This evaluation ensures that the profile's preferences match what would be expected based on its demographic context.
\end{itemize}

\paragraph{Seed Data Details. } 

\input{tables/seed_data_profile}

\input{tables/seed_data_env}

\input{tables/seed_data_shopping}

We provide examples of seed data used in constructing the \texttt{ToolSpectrum}, specifically for building the profile and environment databases in Sec \S\ref{sec:profile and environemnt collection} and for constructing instructions in Sec \S\ref{Instruction Generation} (using the shopping domain as an example). These are illustrated in Table \ref{tab:seed-data-env} and Table \ref{tab:seed-data-example}, respectively.

\paragraph{}
\input{tables/app_list}

\section{Experimental Details}

\subsection{Details of Main Experiments}
In the main experiments, we also test other models, such as LLaMA2-7B-Instruct, Mistral-7B-Instruct-v0.2, and Vicuna-13B-v1.5 \cite{vicuna2023}. However, these models fail to produce results in the correct format, making it difficult to assess their performance accurately.

In addition, we generate experimental code by combining GPT-4O generation with manual modification. We implement it using PyTorch, Transformers, and VLLM open-source packages.

\subsection{The Effects of Different Prompt Methods}\label{context learning}
\input{tables/prompt_method}
Besides simply zero-shot prompting used in the main experiments, we also explore the effects of in-context learning and Chain-of-Thought prompting \cite{kojima2022large, wei2022chain}. Table \ref{tab: diffent-prompt-gpt35} shows the performance of GPT-3.5-turbo and GPT-4o when using different prompt methods. In detail, from the APP perspective, adding shots causes the model's output distribution to be influenced by the examples, preventing it from selecting the appropriate APP based on user profiles and other personalized information. This leads to a decrease in the F1 score. In contrast, including Chain-of-Thought (CoT) enables the model to demonstrate its reasoning process, resulting in an improved F1 score. 
The API undergoes minimal change, as the inherent complexity of API calls and their dependencies remain largely unaffected by the addition of shots or reasoning processes.
Both RP and OP show some improvement, as adding shots helped the model identify the relationship between personalized features and OP, thereby enhancing recall.

\subsection{The Effects of Hierarchical and Flat Prompt} \label{prompting}

\begin{figure}[h!]
    \centering
    \includegraphics[width=\columnwidth]{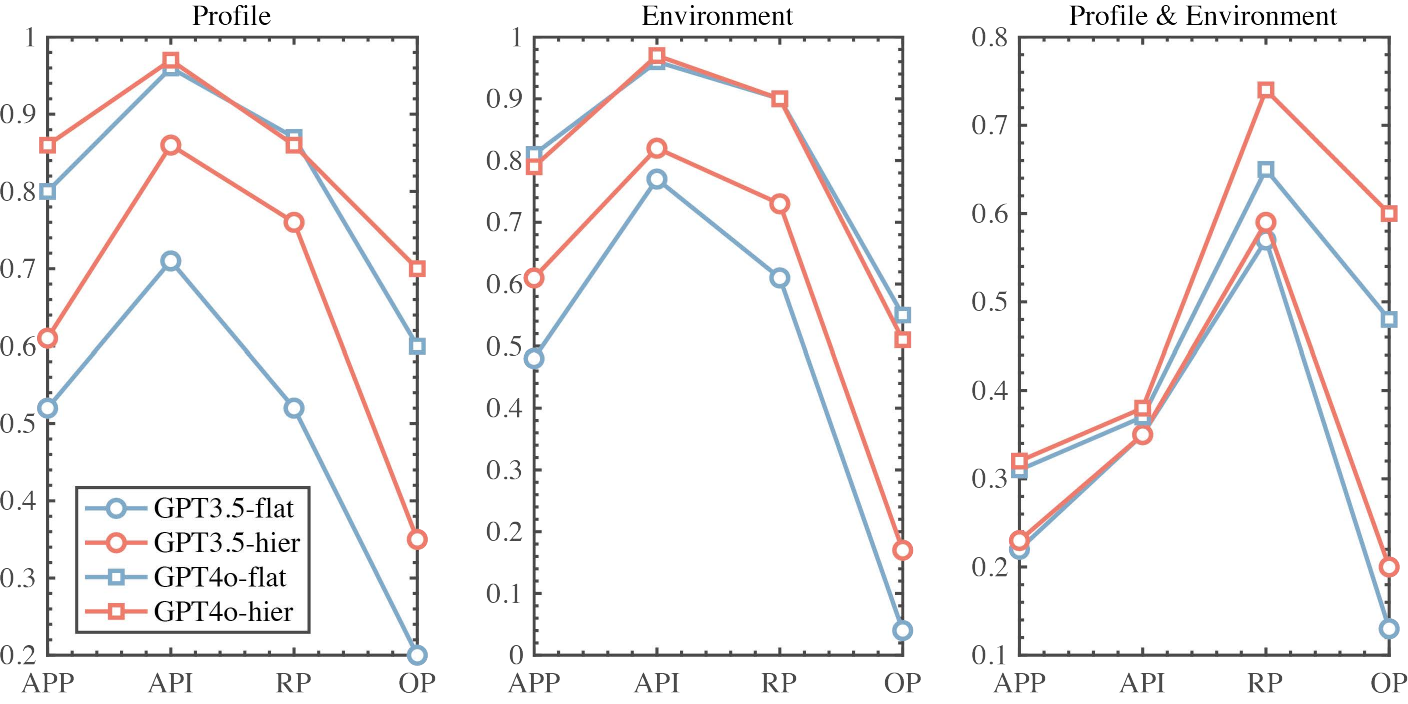}
    \caption{The performance gap between hierarchical and flat prompting on GPT-3.5 and GPT-4o.}
    \label{fig:RQ5}
\end{figure}

In Sec \ref{context learning}, we observe that the effectiveness of Cot was promising when applied in a single-step manner. However, this single-step approach may limit the model’s performance, particularly when dealing with long input contexts that exceed the model's processing capacity. In this section (5.5), we explore a potential improvement: breaking down the Cot process into multiple steps. This hierarchical method involves the model first predicting the relevant domain based on the user's instruction. If the prediction is accurate, the corresponding toolset for that domain is then provided to the model. By doing so, we aim to address the limitations of excessively long contexts and enhance the model's ability to retrieve and process accurate information.

The improvements in all four evaluation dimensions can be observed in Figure \ref{fig:RQ5} when using the hierarchical prompt. This enhancement is primarily due to the reduction of irrelevant toolset information in the context, which effectively lowers noise intensity and allows the model to better understand the relationships between different parameters and personalized information.

More specifically, the performance improvement is greater for GPT-3.5 than for GPT-4o. This is because GPT-4o has a stronger ability to process contextual information and handle noise, making the removal of noise less impactful on its performance. However, in the Profile \& Environment dimension, GPT-4o exhibits a noticeable improvement. By eliminating irrelevant noise, GPT-4o can focus more on understanding the relationship between policy and profile, thereby enhancing its performance. In contrast, GPT-3.5 struggles to comprehend this relationship even after noise removal, resulting in minimal improvement.

\begin{figure}[!htbp]
    \centering
    \includegraphics[width=0.85\linewidth]{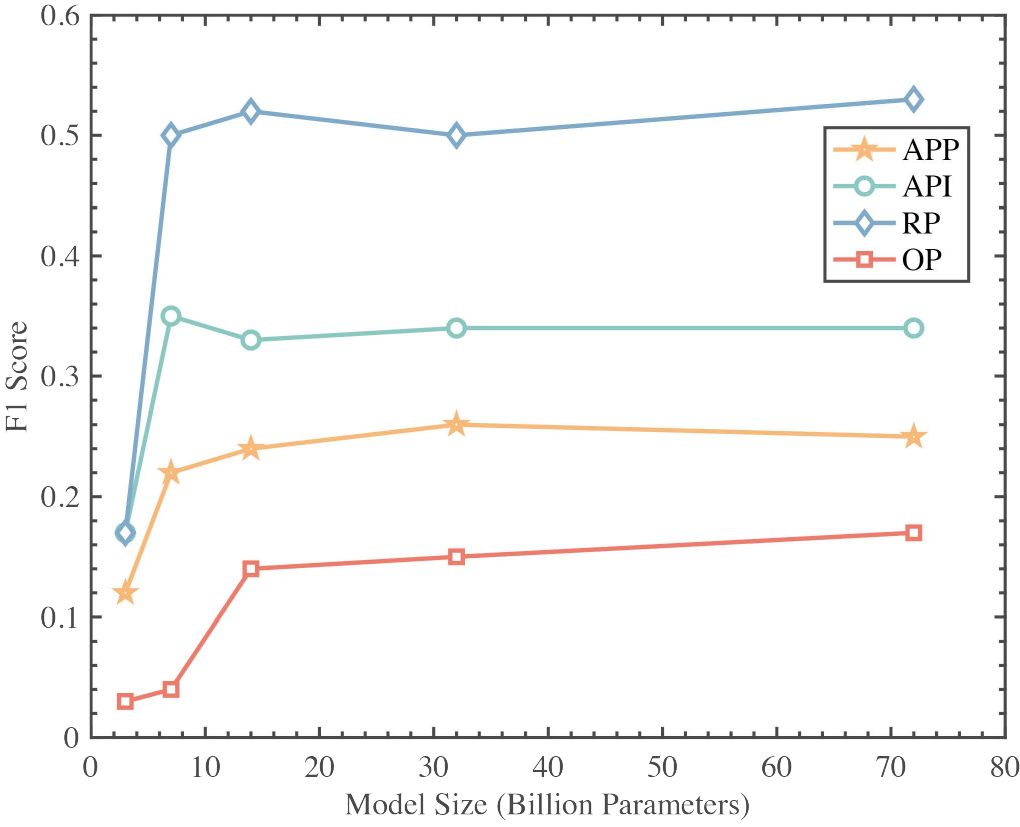}
    \caption{The relationship between the model parameters and performance of the Qwen2.5 series model under the profile\&Environment setting.}
    \label{fig:size_performance}
\end{figure}

\subsection{Cohen's Kappa}
Cohen's Kappa ($\kappa$) measures agreement between two raters classifying items into categories. It quantifies agreement beyond chance, where 1 is perfect, 0 is chance level, and negative values are worse than chance.

Calculated as: $\kappa = \frac{p_o - p_e}{1 - p_e}$, where $p_o$ is observed agreement and $p_e$ is expected agreement due to chance.  $p_o$ is calculated from the confusion matrix diagonal and $p_e$ from marginal probabilities.

Interpretation guidelines (context-dependent): 0.81-1.00 almost perfect, 0.61-0.80 substantial, 0.41-0.60 moderate, 0.21-0.40 fair, 0.00-0.20 slight, < 0.00 poor. These are guidelines only.

\subsection{Error Analysis} \label{sec: error}
\begin{figure*}
    \centering
    \includegraphics[width=\textwidth]{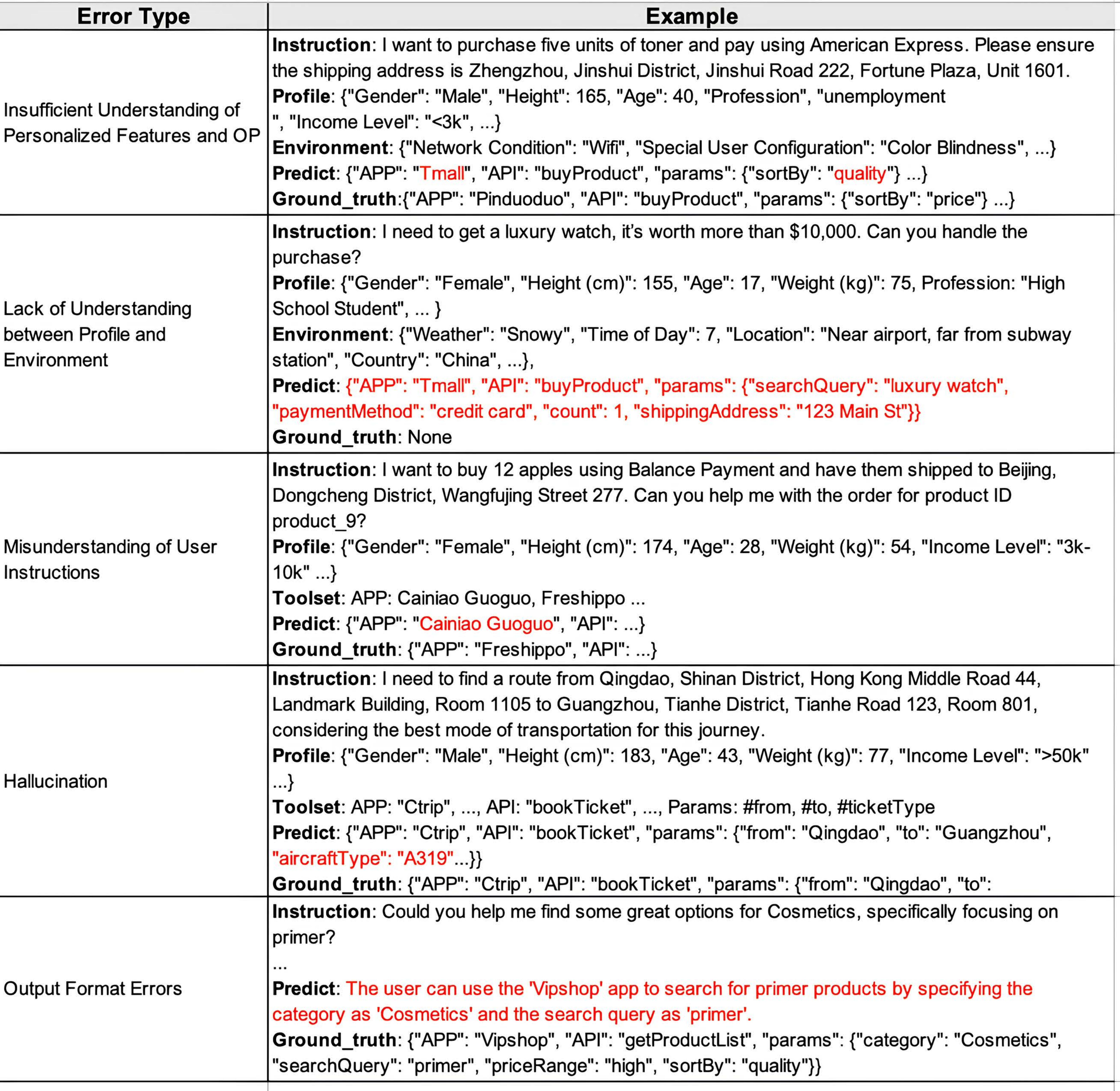}
    \caption{Error examples and specific errors are highlighted in \textcolor{Red}{red}.}
    \label{fig:error-example}
\end{figure*}

\paragraph{\emph{Hallucination (7\%)}}
The hallucination issue remains a significant challenge for large language models \cite{huang2024survey}. The model may fabricate parameters, randomly alter parameter names, or invent function values when generating output, leading to results that do not align with actual logic.

\paragraph{\emph{Output Format Errors (3\%)}}
\texttt{ToolSpectrum} enforces JSON formatting and employs multi-round post-processing, yet the model still struggles to consistently ensure accurate output formats, highlighting the inherent limitations of large models in generating structured data.

\subsection{Case Study}
In this section, as Figure \ref{fig:error-example}, we present examples of various errors, with the specific details highlighted in \textcolor{Red}{red}. \textit{(1) Insufficient Understanding of Personalized Features} and OP: The user is unemployed with a monthly income of less than 3,000. When recommending a shopping app, cost-effectiveness should be prioritized. However, Tmall, known for its high-quality, high-priced products, was suggested, and the recommendation system erroneously prioritized quality when sorting. \textit{(2) Lack of Understanding between Profile and Environment.} In the shopping domain policy, users under the age of 10 are not allowed to purchase items exceeding \$10,000. In this case, the system should return None rather than a result. \textit{(3) Misunderstanding of User Instructions.} The user intended to purchase 12 apples, which should have triggered an action in the grocery domain app. However, the system mistakenly identified the user's intent and invoked Cainiao GuoGuo, a delivery service app. \textit{(4) Hallucination.} The toolset does not contain the \textit{\#aircraftType} parameter, but the model erroneously returned a result with this non-existent parameter due to hallucination. \textit{(5) Output Format Errors}. The model’s response was not in JSON format, which does not comply with the specified format provided in this paper.

\section{Prompt Templates}
\begin{figure*}[!htbp]
    \centering
    \begin{prompt}{Prompt Template for Generating Consumption Preference}

Your current task is to generate a natural language description based on the provided purchase behaviour preference keyword. The description should explain the user's buying habits, preferences, or decision-making style in a realistic and specific way.\\
\\
Example:\\
Keyword: Cost-effectiveness\\
Output: Prefers to buy cost-effective products and is highly sensitive to prices.\\
\\
Keyword: Promotions\\
Output: Frequently shops during promotional events and enjoys comparing prices to find the best deals.\\
\\
Keyword: Eco-certification\\
Output: Prefers purchasing eco-friendly and health-conscious products, paying close attention to eco-certifications.\\
\\
Keyword: {\color{blue}\{keyword\}}\\
Output:

\end{prompt}
    \caption{Prompt Template for Generating Consumption Preference.
    \label{fig: prompt-consumption-preference}
    }
\end{figure*}

\begin{figure*}[!htbp]
    \centering
    \begin{prompt}{Prompt Template for Generating Use Habit}

Your current task is to describe a user's app usage habits based on the provided keyword. The description should detail their typical behavior and preferences when interacting with mobile or web applications.\\
\\
Example:\\
Keyword: Social media\\
Output: Spends at least 1 hour daily on social media platforms, scrolling through feeds in the morning and evening.\\
\\
Keyword: Short video platforms\\
Output: Frequently uses short video platforms during free time to watch entertaining and engaging content.\\
\\
Keyword: Knowledge videos\\
Output: Regularly watch educational content on platforms like YouTube to expand their knowledge base.\\
\\
Keyword: {\color{blue}\{keyword\}}\\
Output:

\end{prompt}
    \caption{Prompt Template for Generating Use Habit.
    \label{fig: prompt-use-habit}
    }
\end{figure*}

\begin{figure*}[!htbp]
    \centering
    \begin{prompt}{Prompt Template for Generating Content Preference}

Your current task is to write a description of a user's content consumption habits based on the provided keyword. The description should detail their interests and patterns when engaging with online content.\\
\\
Example:\\
Keyword: Short videos\\
Output: Enjoys watching short videos and live streams, especially those focusing on entertainment and humor.\\
\\
Keyword: Tech articles\\
Output: Prefers reading technology and finance-related articles and staying informed about industry trends.\\
\\
Keyword: Travel and food\\
Output: Loves exploring travel guides and food content, often seeking inspiration for their next adventure.\\
\\
~Keyword: {\color{blue}\{keyword\}}\\
Output:

\end{prompt}
    \caption{Prompt Template for Generating Content Preference.
    \label{fig: prmopt-content-preference}
    }
\end{figure*}

\begin{figure*}[!htbp]
    \centering
    \begin{prompt}{Prompt Template for Scoring Profile}

I am currently grading generated virtual personas based on their features' internal consistency and plausibility.\\
\\
Requirements:\\
\\
1.  The profile must include the following attributes: gender, height, age, weight, occupation, income level, and preferences (including consumption preferences, content consumption preferences, and app usage history preferences).\\
2.  The score reflects the [plausibility] and [internal consistency] of the profile. This means:\\
    - Demographic Coherence: The combination of age, gender, height, weight, occupation, and income should form a believable profile. For example, a 20-year-old high school dropout is unlikely to have a high income. Similarly, a very short, elderly person is unlikely to be a professional basketball player.\\
    - Preference Alignment: The profile's preferences (consumption, content, app usage) should align with their demographics, especially in terms of consumption preferences and income levels. For example, Individuals with an income level below 10k are less likely to frequently purchase high-quality goods, while those with an income above 50k typically place greater emphasis on the quality of the goods. A retired individual is less likely to use professional networking apps heavily.\\
3.  The scoring range is integer values from 1 to 10, where 1 is highly implausible and 10 is perfectly plausible.\\
4.  You must return only an integer value (1-10) and no other output.\\
\\
Input:
{\color{blue} \{profile\}}\\
\\
Output:

\end{prompt}
    \caption{Prompt Template for Scoring Profile.
        \label{prompt-scoring-profile}
    }
\end{figure*}

\begin{figure*}[!htbp]
    \centering
    \begin{prompt}{Prompt Template for Generating Instruction}

I am creating a dataset for tool calls, and I will input some keywords for you. Please help me output the corresponding user commands based on the keywords.\\
\\
requirement:\\
1. The input includes an API description and keywords. The API description is an introduction to the current application scenario; Keywords are a dictionary containing some keywords, and you need to concatenate them into instructions that fit the user's usage habits.\\
2. The generated instruction is in the first person.\\
3. The generated instructions must include these keywords and cannot be modified in any way.\\
4. Do not output any irrelevant content, instructions should not contain any irrelevant symbols, and only natural language is allowed.\\
5. Do not include keys in the keyword dictionary, as the generated instructions can only contain values.\\
6. Do not explicitly express the content of the API Description.\\
\\
\#\#\# Input:\\
\#\#API Description\\
{\color{blue}\{api description\}}\\
\\
\#\# Keywords:\\
{\color{blue}\{keywords\}}\\
\\
\#\#\# Output:\\

\end{prompt}
    \caption{Prompt Template for Generating Instruction.
    \label{prompt-instruction}
    }
\end{figure*}

\begin{figure*}[!htbp]
    \centering
    \begin{prompt}{Prompt Template for Generating Tool-call Results}

You are a personalized tool assistant. I will provide you with user instructions, a user profile, an external environment, and a toolset. Please output the most reasonable tool invocation result.\\
\\
Requirements:\\
\\
1. The input consists of user instructions, user profile, and the external environment. The user instruction is a string that describes the user's needs; the user profile is a dictionary that includes the user's gender, height, age, weight, occupation, income level, and preferences (including consumption preferences, content consumption preferences, and app usage history preferences); the environment is a dictionary that includes Weather, Time of Day, Location, Date (month-day), and Network Condition; the toolset is a dictionary that includes a description of each app's functionality, along with all APIs and their corresponding parameters, parameter types, and parameter ranges.\\
2. You should comprehensively consider the user instruction, profile, and environment to give the most appropriate tool invocation result.\\
3. The output format should be a dictionary containing the app name, API name, corresponding parameters, and the explanation, as shown in the following format:\\
    \{\\
        \ \ \ \ "APP": "APP",\\
        \ \ \ \ "API": "API",\\
        \ \ \ \ "params": \{\\
            \ \ \ \ \ \ \ \ "key1": "value1",\\
            \ \ \ \ \ \ \ \ "key2": "value2",\\
            \ \ \ \ \ \ \ \ ...\\
        \ \ \ \ \}\\
        \ \ \ \ "explanation": "EXPLANATION"\\
    \}\\

4. There will be a policy field in the environment, and if you feel that the user's instruction violates the user's profile, set the APP, API, and params to null.\\
5. Only output one dictionary, do not output any other content.\\
\\
\\
\#\#\# Input\\
\#\# User Instruction\\
{\color{blue}\{instruction\}}\\
\\
\#\# User profile\\
{\color{blue}\{profile\}}\\
\\
\#\# Environment\\
{\color{blue}\{environment\}}\\
\\
\#\# Toolset\\
{\color{blue}\{toolset\}}\\
\\
\#\#\# Output\\
\\

\end{prompt}
    \caption{Prompt Template for Generating Tool-call Results.
    \label{prompt-call-result}
    }
\end{figure*}

\begin{figure*}[!htbp]
    \centering
    \begin{prompt}{Prompt Template for Scoring Tool-call Result}

I am currently evaluating the quality of tool-calling datasets. I will provide you with user instructions, user profile, external environment, and tool calling results. Please comprehensively consider these factors and tell me whether the tool's calling results meet expectations.\\
\\
Requirements:\\
\\
1. Please judge whether the tool calling results meet expectations based on user instructions, user profile, external environment, and tool calling results.\\
2. Please rate the following three aspects separately, ranging from 1 to 10 points. The output format is separated by commas.\\
3. You should evaluate whether the tool calling results meet expectations from three aspects:\\
    - Whether the tool calling results can solve the user’s needs;\\
    - Whether the tool calling results match the user’s profile;\\
    - Whether the tool calling results align with the external environment.\\
4. Do not output any irrelevant content, only output the answer.\\
\\
Below are the user instructions, user profile, external environment, and tool calling results I’m inputting:\\
\\
\\
\#\#\# Input\\
\#\# User Instruction\\
{\color{blue}\{instruction\}}\\
\\
\#\# User profile\\
{\color{blue}\{profile\}}\\
\\
\#\# Environment\\
{\color{blue}\{environment\}}\\
\\
\#\# Tool Calling Result\\
{\color{blue}\{tool call result\}}\\
\\
\#\#\# Output\\
\\

\end{prompt}
    \caption{Prompt Template for Scoring Tool-call Result.
    \label{prompt-qulity-call-result}
    }
\end{figure*}

\begin{figure*}[!htbp]
    \centering
    \begin{prompt}{Prompt Template for Selecting the Superior Results}

I am conducting an evaluation of personalized tool utilization. Analyze which response better fulfills user needs by following these requirements:\\
\\
requirement:\\
\\
1. The input includes a description of the toolset, user instructions, and results A and B.\\
2. You need to determine which result better meets the user's needs.\\
3. You can only output A or B, do not output any other irrelevant content.\\
\\
\#\#\# Input\\
\#\# Toolset\\
{\color{blue}\{toolset\}}\\
\\
\#\# User Profile\\
{\color{blue}\{profile\}}\\
\\
\#\# Response A\\
{\color{blue}\{response\_A\}}\\
\\
\#\# Response B\\
\\
{\color{blue}\{response\_B\}}\\
\\
\#\#\# Output\\

\end{prompt}
    \caption{Prompt Template for Selecting the Superior Results.
    \label{fig:prompt-select}
    }
\end{figure*}

%% file: tables/seed_data_profile.tex
\begin{table*}[]
\centering
\begin{tabular}{ll}
\toprule[2pt]
\multicolumn{1}{c}{\textbf{Parameter}} & \multicolumn{1}{c}{\textbf{Option}}                                              \\ \midrule[1pt]
Gender                                 & Male, Female                                                                     \\
Age                                    & \multirow{3}{*}{No need for seed data, sample directly as a normal distribution} \\
Height                                 &                                                                                  \\
Weight                                 &                                                                                  \\
Profession                             & Designer, Student, Teacher, Doctor, Farmer                                       \\
Income Level                           & <3k, 3k-10k, 10k-50k, >50k                                                       \\
Education Background                   & Bachelor, Master, PhD, ...                                                       \\
Personality \& Preferences &
  \begin{tabular}[c]{@{}l@{}}Cost-effectiveness, Product quality, service experience\\ Science discoveries,  DIY and crafts,  Gaming\\ Tmall, Ctrip, Vishop, Amazon, ...\end{tabular} \\ \bottomrule[2pt]
\end{tabular}
\caption{Seed data for Profile database collection, used for subsequent sampling and generation of Profile database $\mathcal{P}$.}
\label{tab:seed-data-profile}
\end{table*}

%% file: tables/seed_data_env.tex
\begin{table*}[]
\centering
\begin{tabular}{ll}
\toprule[2pt]
\multicolumn{1}{c}{\textbf{Parameter}} & \multicolumn{1}{c}{\textbf{Option}}                 \\ \midrule[1pt]
Weather                                & Sunny, Rainy, Snowy, Thunderstorm, ...              \\
Time                                   & 0:00, 1:00, 2:00, 3:00, ...                         \\
Date                                   & 01-01, 01-02, 01-03, ...                            \\
Location                               & China, USA, India, Germany, France, ...             \\
Network Condition                      & Wifi, Mobile Network, No Network, ...               \\
Battery Level                          & 100\%, 99\%, 98\%, 97\%, ...                        \\
Device-specific Configuration          & Hearing Impairment, Blindness, Color Blindness, ... \\
\bottomrule[2pt]
\end{tabular}%
\caption{Seed data for Environment database collection, used for subsequent sampling and generation of Environment database $\mathcal{E}$.}
\label{tab:seed-data-env}
\end{table*}

%% file: tables/seed_data_shopping.tex
\begin{table*}[]
\centering
\begin{tabular}{ll}
\toprule[2pt]
\multicolumn{1}{c}{\textbf{Parameter}} & \multicolumn{1}{c}{\textbf{Option}} \\ \midrule[1pt]
\multirow{7}{*}{Search keyword} & \textbf{Clothing}: dress, shoes, t-shirt, jeans, coat, skirt, socks, ... \\
 & \textbf{Electronics}: laptop, phone, tablet, camera, headphones, ... \\
 & \textbf{Furniture}: sofa, table, chair, bed, desk, bookshelf, ... \\
 & \textbf{Books}: fiction book, textbook, novel, biography, ... \\
 & \textbf{Toys}: toy car, doll, puzzle, board game, ... \\
 & \textbf{Food}: snack, chocolate, candy, instant noodles, ... \\
 & \textbf{Sports}: basketball, football, tennis racket, yoga mat, ..., \\
Conditions & brand new, slightly used, heavily used \\
Shipping addresses & Beijing, London,  New York, ... \\
Payment methods & WeChat Pay, Alipay, UnionPay, ... \\ \bottomrule[2pt]
\end{tabular}
\caption{Seed data for \textit{Shopping} domain, used for subsequent sampling and generation of user instructions $\mathcal{I}$.}
\label{tab:seed-data-example}
\end{table*}

%% file: tables/app_list.tex
\begin{table*}[!htbp]
\begin{adjustbox}{width=\textwidth}

\begin{tabular}{ccc}
\toprule[2pt]
\textbf{Domain} & \textbf{APPs}                               & \textbf{APIs}                                                     \\ \midrule[1pt]
Shopping        & Temu, Amazon, Poizon, Vipshop, Xianyu & \textit{getProductList, buyProduct}                               \\
Travel & Baidu\_Maps, Didi\_Chuxing, Ctrip & \textit{getDistance, getRoute, bookTaxi, rentCar, bookTicket, bookHotel} \\
Entertainment   & Maoyan, Damai                              & \textit{getShowSchedule, bookShowTicket}                         \\
Grocery         & Freshippo, Duoduo Maicai                   & \textit{getProductList, buyProduct}                              \\
Delivery        & Cainiao Guoguo                             & \textit{createShipment, getShipmentStatus, getCourierLocations,} \\
Finance         & Bank, Tonghuashun                          & \textit{getFundDetails, buyFund, getStockDetails, buyStock}      \\
Health          & Ping An Health, Keep                       & \textit{createHealthPlan, logExercise}                           \\
Knowledge       & Xiaohongshu, Zhihu, Dedao                  & \textit{getKnowledge}                                            \\
News            & Toutiao, Weibo, Hupu                       & \textit{getDailyNewsRecommendations}                             \\ \bottomrule[2pt]
\end{tabular}
\end{adjustbox}
\caption{List of all Apps and their corresponding APIs in the \texttt{ToolSpectrum}.}
\label{tab:app_list}
\end{table*}

%% file: tables/prompt_method.tex
\begin{table*}[h!]
\centering
\begin{adjustbox}{width=0.85\textwidth}
\begin{tabular}{llcccc|cccc|cccc}
\toprule[2pt]
\multirow{2}{*}{\textbf{Model}} & \multicolumn{1}{c}{\multirow{2}{*}{\textbf{Method}}} & \multicolumn{4}{c|}{\textbf{Profile}} & \multicolumn{4}{c|}{\textbf{Environment}} & \multicolumn{4}{c}{\textbf{Profile \& Environment}} \\ \cline{3-14} 
 & \multicolumn{1}{c}{} & APP & API & RP & OP & APP & API & RP & OP & APP & API & RP & OP \\ \midrule[1pt]
\multirow{3}{*}{GPT-3.5-turbo} & Zero shot & 0.52 & \textbf{0.50} & 0.40 & 0.15 & 0.48 & 0.47 & 0.35 & 0.04 & 0.22 & 0.12 & 0.34 & 0.02 \\
 & Few shot & 0.37 & \textbf{0.50} & 0.48 & 0.26 & 0.27 & \textbf{0.51} & 0.35 & 0.23 & 0.23 & 0.08 & 0.33 & 0.04 \\
 & CoT & \textbf{0.58} & 0.49 & \textbf{0.54} & \textbf{0.34} & \textbf{0.51} & 0.50 & \textbf{0.36} & \textbf{0.25} & \textbf{0.24} & \textbf{0.17} & \textbf{0.37} & \textbf{0.07} \\ \midrule[1pt]
\multirow{3}{*}{GPT-4o} & Zero shot & 0.80 & 0.77 & \textbf{0.73} & 0.52 & \textbf{0.81} & 0.80 & 0.74 & 0.50 & 0.32 & \textbf{0.30} & 0.62 & 0.45 \\
 & Few shot & 0.75 & \textbf{0.78} & 0.72 & 0.55 & 0.77 & \textbf{0.81} & 0.74 & 0.68 & 0.32 & 0.18 & 0.63 & 0.46 \\
 & CoT & \textbf{0.82} & 0.76 & \textbf{0.73} & \textbf{0.56} & 0.75 & 0.73 & \textbf{0.75} & \textbf{0.73} & \textbf{0.34} & 0.25 & \textbf{0.64} & \textbf{0.49} \\ \bottomrule[2pt]
\end{tabular}
\end{adjustbox}
\caption{Performance of GPT-3.5-turbo and GPT-4o based on different prompt methods.}
\label{tab: diffent-prompt-gpt35}
\end{table*}